# A Light weight and Hybrid Deep Learning Model based Online Signature Verification

Chandra Sekhar V
IIIT Sricity
Sricity, India
chandrasekhar.v@iiits.in

Anoushka Doctor
University of Melbourne
Victoria 3010, Australia
doctora@student.unimelb.edu.au

Prerana Mukherjee
IIIT Sricity
Sricity, India
prerana.m@iiits.in

Viswanath P
IIIT SriCity
SriCity, India
viswanath.p@iiits.in

**Abstract**

*The augmented usage of deep learning-based models for various AI related problems are as a result of modern architectures of deeper length and the availability of voluminous interpreted datasets. The models based on these architectures require huge training and storage cost, which makes them inefficient to use in critical applications like online signature verification (OSV) and to deploy in resource constraint devices. As a solution, in this work, our contribution is two-fold. 1) An efficient dimensionality reduction technique, to reduce the number of features to be considered and 2) a state-of-the-art model CNN-LSTM based hybrid architecture for online signature verification. Thorough experiments on the publicly available datasets MCYT, SUSIG, SVC confirms that the proposed model achieves better accuracy even with as low as one training sample. The proposed models yield state-of-the-art performance in various categories of all the three datasets.*

## 1. Introduction

Online Signature Verification (OSV) is a challenging research problem in the area of Artificial Intelligence. OSV finds numerous critical applications like e-signatures, banking transaction, online financial transactions [1,2,4,6] etc. Recent advancements in mobile networking/ devices and touchscreen technology lead to utilization of specialized interfaces to collect the unique signing characteristics like pressure, angle of stylus pen, velocity etc., at each point of signature and analyzing these key features along with the geometric co-ordinates. This numerical information is used for online signature verification in contrast to a static image verification for offline signature system [1,4,5,7,15,16,43].

In literature[1,6-9,20], the online signature verification models are broadly classified into two types 1) Traditional feature extraction based 2) Deep learning technique based verification systems. In traditional feature extraction based OSV models, the authenticity of a test signature is decided by means of a appropriate matching technique based on pattern recognition techniques such as Symbolic classifier[3,7,9], Dynamic Time Warping[13,25,31], Hidden Markov Model [4,6], Support Vector Machine [43], Neuro fuzzy [9,11,12], Random forest[2-4], Neural Networks [3,4], Viterbi path [4], etc.

The recent advancements in computing resources and increasing amount of accessibility to huge datasets leads to evolution of Machine Learning and Deep learning Techniques (MLDL). The advancements in MLDL [28,41,42] techniques lead to development of models of deeper architecture and ability to process huge amount of data. In case of traditional OSV models, Zhang et al [1] proposed a first of its kind of an attempt for OSV based on template matching technique in which the authenticity of a writer is determined by comparing an input signature with a corresponding user reference set. Based on the similarity distance, the signature is classified as genuine or forgery and achieved an Average Error Rate (AER) of 2.2% on a custom dataset. Cpałka et al [10], proposed an OSV model by segmenting the signature into sections called partitions which represents the time moments. A neuro fuzzy classifier is used to classify the signature based on partitions. The model achieved an AER of 10.70%. In their extension work, Cpałka et al [11,12] proposed novel works, in which writer specific partitions of the signatures are selected by eliminating redundant partitions and achieved an EER of 3.24%.

Very few (only two) works have been proposed based on MLDL techniques for OSV. Tolosona et al [45] proposed a novel work in which a Siamese architecture based on Recurrent Neural Networks (RNNs) is proposed to learn a dissimilarity metric from the pairs of signatures. The dissimilarity metric is used to classify the signature and achieved an AER of 6.22%. The second work is by Lai et al [27], in which RNNs are trained to learn a scale invariance and rotation invariance feature called the 'length-normalized path signature', LNPS helps in classifying the signature. Lai et al achieved an EER of 2.37% on SVC-2004 dataset.

Despite the lower error rates, MLDL frameworks need comparatively large number of training samples for each user to learn the inter-individual variability, intra-individual variability [1,4-6,30], to efficiently classify the genuineness of signatures [13,30,31]. Nevertheless, it is often impractical to acquire satisfactory number of signature samples from users, given the sensitivity of applications e.g., m-banking.

In this context, very few works explored the likelihood of OSV systems with few shot learning i.e. learning the user specific features with one/few signature samples. Galbally et al [16] proposed an OSV framework in which synthetic samples are generated from one signature sample by duplicating the signature using Hidden Markov Models. Another work in the same direction is by Diaz et al [24,29], in which, single samplings were duplicated based on the kinematic theory of rapid human movements, and its sigmalognormal parameters. This model achieved an Equal Error rate (EER) of 13.56% with MCYT-100 dataset, when the model is trained with single signature sample.

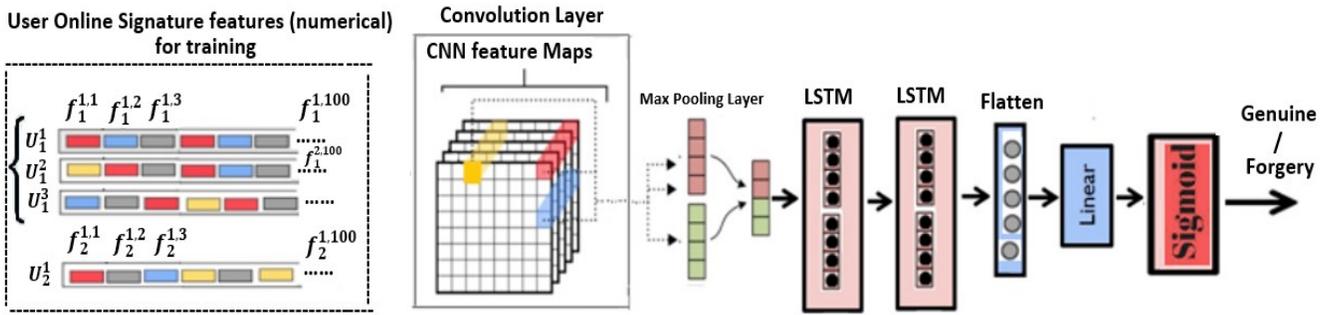

Figure 1: Proposed Deep CNN + LSTM based architecture for Online Signature Verification. In the diagram, Flatten stands for fully connected layers.

## 2. Our Contribution

**The main contributions of our work can be summarized as follows:**

1. In this paper, each user / writer original features from the dataset are independently clustered using traditional K-Means clustering algorithm. The clustering technique results in set of feature clusters. A cluster representative for each cluster is selected and the set of these cluster representatives forms the reduced feature subset for each user. The cluster representatives are selected through statistical dispersion measure: Mean absolute difference (MAD).

2. We present a combination of two deep neural network architectures viz., a Convolutional Neural Network (CNN) and a Long Short-Term Memory (LSTM) in which the reduced feature set (cluster heads) are given as input to the CNN layer. CNN layer extracts the local deep features which are fed into the LSTM layers. The long-term dependencies learnt by LSTM in sequential data are used for signature classification (genuine or forgery).

3. Exhaustive evaluation of the proposed model by conducting experiments on three most widely used datasets for OSV i.e. MCYT-100, SUSIG and SVC.

The manuscript is organized as follows. In Section 3, we present different phases of our proposed model. In section 4, details of training and testing data, experimental analysis along with the results produced by the model are discussed. A comparative analysis of the proposed model with other latest related models is reported in section 5. Conclusions are drawn in section 6.

## 3. Proposed Model Architecture

Inspired by the works from [39,40,41,42] in sequence modelling, sentiment analysis [39], attentiveness [44] and the fact that CNN can extract best local features of input and RNN (recurrent neural network) and its variants (LSTM, RGU) can process sequence input and learn the long-term dependencies, we combined both of them for online signature (which is a sequence of points) verification analysis. As depicted in fig 1, the reduced feature set which are the cluster representatives forms an input to the Convolutional Neural Network layer (CNN). The CNN learns the low-level translation invariant features from the reduced feature set and feed forward to LSTMs (Long Short-Term Memory) layers in order to compose higher order features.

Our model consists of the following components: reducing the original feature set, convolutional and pooling layers, concatenation layer, LSTM layer, fully connected layer with Sigmoid output. These fragments are discussed below:

### 3.1 Writer Dependent Feature Clustering.

let S = $[S_1^j, S_2^j, S_3^j, ..., S_n^j]$ be a set of 'n' signature samples of writer 'j' i.e. $U_j$, j = 1, 2, 3,….N. (N represents the number of writers). Let $F = [F_1^j, F_2^j, F_3^j, ..., F_m^j]$ be a set of m-dimensional combined feature vectors, where $F_i^j = [f_{i1}^j, f_{i2}^j, f_{i3}^j, ..., f_{in}^j]$ be the feature set characterizing the i$^{th}$ feature of signature samples of writer 'j' i.e. $U_j$. The Feature-Signature matrix (FS) of writer $U_j$ is shown below:

| F/S | $S_1$ | $S_2$ | $S_3$ | $S_n$ | Mean($\overline{MF_i}$) |
|---|---|---|---|---|---|
| $F_1$ | $f_{11}$ | $f_{12}$ | | $f_{1n}$ | |
| $F_2$ | | | | | |
| $F_m$ | $f_{m1}$ | $f_{m2}$ | | $f_{mn}$ | |

The feature vectors $F_1, F_2, …, F_m$ forms an input to the K-Means clustering technique in which each feature is grouped under one of the 'K' clusters. For each cluster returned by the clustering technique, we will compute the cluster representative using a novel statistical dispersion metric which is described below.

### 3.2 Computing the Cluster Representatives

In literature [2,7,14], to select a subset of 'd' writer dependent features based on their relevance from a total set of 'P' features, the widely used dispersion measures for selecting writer dependent features are: Mean absolute difference (MAD) =

$$MAD_k = \frac{1}{n} \sum_{i=1}^{n} |f_{ik} - \overline{f_k}|, k = 1,2,3, ..., P \quad (1)$$

where $\overline{f_k}$ is the mean of the k$^{th}$ feature of the writer W$_j$ and $f_{ik}$ is the feature value of the k$^{th}$ feature for the i$^{th}$ sample of the writer W$_j$. $MAD_k$ is computed for each feature of the feature set i.e., $k = 1,2,3,...,P$.

To select one feature among the features of a cluster as a cluster representative, we have computed MAD by applying the formula (2). The feature which is having maximum MAD value among the features of a corresponding cluster is selected as cluster representative. On computing the cluster representative for all the clusters, the set of cluster representatives are grouped to form a reduced feature subset 'FS'. The 'FS' forms an input to the Convolutional Neural Network. In case of MCYT-100 dataset, as shown in Table II, the features are reduced from 100 to 80 and 100 to 50. Each signature is a one – dimensional vector of length 50 i.e. 1* 50 in case of 50 features, 1* 80 in case of 80 features.

### 3.3 Convolution and Pooling

On receiving the reduced feature set 1*80, the Convolutional Neural Network performs a one-dimensional convolution operation (matrix-vector multiplication) between the two signals i.e. input sequence vector and the sliding kernel (weight matrix) to extract local features for each window of the given signature. Sliding the weight matrix all over the signature feature sequence generates the advanced features, consolidating these features produce Feature Maps. In our implementation, we have used 32 filters, each of size 5 (one dimensional) which results in an output of a feature vector of size 80 × 36. On the output of first conv layer, batch normalization is applied to regulate the input to the activation function and for faster convergence, which results in an output of 80 × 36. Likewise, the second convolution layer uses 64 filters of size 1 ×3 and outcomes a feature vector of size 80 ×36. These feature representations form an input to the LSTM layer of 34 units.

### 3.4 Long Short Term Memory (LSTM)

LSTMs are explicitly designed with a default behavior to process sequential input and to learn the long-term dependencies and are well-suited to classify temporal data given time lags of unknown duration. The LSTMs have an exceptional property of relative insensitivity to input gap length, in which the other models like RNNs, hidden markov models and other sequence learning methods fails [39,40,41,42]. As online signature is a sequence of time series points, we have used LSTM layers to learn long term dependencies of signature features. LSTM layer outputs a feature vector of size 80 × 32. The deep representational features from the LSTM layers is given as an input to the fully connected layers.

### 3.5 Fully Connected Network with Sigmoid Output

The proposed framework uses a two hidden layered Multilayer Perceptron (MLP) as classifier. A deep feature vector from the LSTM layer of size 80 × 32 forms an input to the first fully connected layer of MLP. Flatten reshapes the feature vector of size 80 × 32 into a high-level feature representation of size 1×2560. The number of neurons in the first fully connected layers are 32. The output of size 1×32 from the first fully connected layer is given as input to the second fully connected layer which contains 32 neurons. A feature vector of size 1×32 from the second fully connected layers forms an input to the batch normalization and dropout layers which outputs a feature of size 1×32. The final feature forms an input to final sigmoid layer for final classification into genuine or forgery. We have used 'binary_crossentropy' as the loss function that measures the discrepancy between the real signature class and the model output and 'adam' as an optimizer, with batch size of 16 and total of 800 epochs.

TABLE I. THE DATASET DETAILS USED IN THE EXPERIMENTS FOR THE PROPOSED MODEL

| DataSet → | MCYT-100 | SVC | SUSIG |
|---|---|---|---|
| # of Users | 100 | 40 | 94 |
| Total Number of features | 100 | 47 | 47 |
| Training (Genuine+Training) | 3600 (72%) | 1120 (70%) | 1880 (67%) |
| Testing (Genuine) – FRR | 700 (14%) | 240 (15%) | 564 (20%) |
| Testing (Forgery) - FAR | 700 (14%) | 240 (15%) | 376 (13%) |
| Total Testing Samples % | 28% | 30% | 33% |
| Total Number of Samples | 5000 | 1600 | 2820 |

TABLE II. COMPARATIVE ANALYSIS OF THE PROPOSED MODEL AGAINST THE RECENT MODELS ON MCYT (DB1) DATABASE (where 'S' and 'R' represents Skilled and Random categories respectively. The number indicates the number of signature samples used for training).

| Method | S_01 | S_05 | S_10 | S_15 | S_20 | R_01 | R_05 | R_10 | R_15 | R_20 |
|---|---|---|---|---|---|---|---|---|---|---|
| **Proposed Model – (Hybrid Deep Learning Model + few shot learning)** | 15.57 | 1.88 | 0.67* | 0.73* | 0.00* | 16.70 | 0.16 | 0.04* | 0.06* | 0.00* |
| GMM+DTW with Fusion [16] | - | 3.05 | - | - | - | - | - | - | - | - |
| Cancelable templates - HMM Protected [17] | - | 10.29 | - | - | - | - | - | - | - | - |
| Cancelable templates - HMM [17] | - | 13.30 | - | - | - | - | - | - | - | - |
| Histogram + Manhattan [20] | - | 4.02 | - | - | - | - | 1.15 | - | - | - |
| discriminative feature vector + several histograms [20] | - | 4.02 | - | - | 2.72 | - | 1.15 | - | - | 0.35 |

| Method | | | | | | | | | | |
|---|---|---|---|---|---|---|---|---|---|---|
| Writer dependent parameters (Symbolic) [21] | - | 2.2 | - | - | 0.6 | - | 1.0 | - | - | 0.1** |
| VQ+DTW[25] | - | 1.55* | - | - | - | - | - | - | - | - |
| writer dependent features and classifiers[26] | - | 19.4 | - | - | 1.1 | - | 7.8 | - | - | 0.8 |
| Stroke-Wise [29] | 13.72** | - | - | - | - | 5.04* | - | - | - | - |
| Target-Wise [29] | 13.56* | - | - | - | - | 4.04** | - | - | - | - |
| Information Divergence-Based Matching [30] | - | 3.16 | - | - | - | - | - | - | - | - |
| WP+BL DTW[31] | - | 2.76 | - | - | - | - | - | - | - | - |
| Representation learning + DTW (Skilled forgery) [34] | | 1.62** | | | | | 0.23 | | | |
| Representation learning + DTW (Random forgery) [34] | | 1.81 | | | | | 0.24 | | | |
| Combinational Features and Secure KNN-Global features [35] | - | 5.15 | - | - | - | - | 1.70 | - | - | - |
| Combinational Features and Secure KNN-Regional features [35] | - | 4.65 | - | - | - | - | 1.33 | - | - | - |
| Stability Modulated Dynamic Time Warping (F13) [35] | - | 13.56 | - | - | - | - | 4.31 | - | - | - |
| Dynamic Time Warping-Normalization(F13) [35] | - | 8.36 | - | - | - | - | 6.25 | - | - | - |
| Writer dependent parameters (IntervalValued representation) [36] | - | 2.51 | - | - | 0.03** | - | 0.70 | - | - | 0.00* |
| Common feature dimension and threshold (IntervalValued representation) [36] | - | 10.36 | - | - | 5.82 | - | 10.32 | - | - | 0.74 |
| Writer dependent parameters (conventional) [36] | - | 6.79 | - | - | 0.00* | - | 1.73 | - | - | 0.00* |
| Common feature dimension and threshold (conventional) [36] | - | 13.12 | - | - | 11.23 | - | 5.61 | - | - | 1.66 |
| Probabilistic-DTW(case 1) [37] | - | - | - | - | | - | 0.0118* | - | - | - |
| Probabilistic-DTW(case 2) [37] | - | - | - | - | - | - | 0.0187** | - | - | - |
| Curvature feature [38] | - | 10.22 | 8.25 | 6.38 | - | - | 4.12 | 3.33 | 2.58 | - |
| Torsion Feature [38] | - | 9.22 | 7.04 | 5.12 | - | - | 3.42 | 2.25 | 1.90 | - |
| Curvature feature +Torsion Feature[38] | - | 6.05 | 4.23** | 3.10** | - | - | 2.95 | 1.81** | 1.20** | - |

TABLE III. COMPARATIVE ANALYSIS OF THE PROPOSED MODEL AGAINST THE RECENT MODELS ON SVC DATASET

| Method | S_01 | S_05 | S_10 | S_15 | R_01 | R_05 | R_10 | R_15 |
|---|---|---|---|---|---|---|---|---|
| **Proposed Model – (Hybrid Deep Learning Model + few shot learning)** | 6.71* | 1.05* | 0.00* | 0.10* | 9.53 | 0.16 | 0.18* | 0.16* |
| LCSS (User Threshold) [19] | - | - | 5.33 | - | - | - | - | - |
| RNN+LNPS[27] | - | - | - | - | - | 2.37 | - | - |
| Target-Wise [29] | 18.63 | - | - | - | 0.50* | - | - | - |
| Stroke-Wise [29] | 18.25** | - | - | - | 1.90** | - | - | - |
| DTW based (Common Threshold) [31] | - | - | 7.80 | - | - | - | - | - |
| Stroke Point Warping [32] | - | - | 1.00** | - | - | - | - | - |
| SPW+mRMR+SVM(10-Samples) [32] | - | - | 1.00** | - | - | - | - | - |
| Variance selection [33] | - | - | 13.75 | - | - | - | - | - |
| PCA [33] | - | - | 7.05 | - | - | - | - | - |
| Relief-1 (using the combined features set) [33] | - | - | 8.1 | - | - | - | - | - |
| Relief-2 [33] | - | - | 5.31 | - | - | - | - | - |
| Probabilistic-DTW(case 1) [37] | - | - | - | - | - | 0.0025* | - | - |
| Probabilistic-DTW(case 2) [37] | - | - | - | - | - | 0.0175** | - | - |
| Curvature feature +Torsion Feature[38] | - | 9.83** | 6.61 | 3.10** | - | 3.54 | 1.24** | 1.81** |

TABLE IV. COMPARATIVE ANALYSIS OF THE PROPOSED MODEL AGAINST THE RECENT MODELS ON SUSIG DATASET

| Method | S_01 | S_05 | S_10 | R_01 | R_05 | R_10 | Number of Samples for training |
|---|---|---|---|---|---|---|---|
| **Proposed Model – (Hybrid Deep Learning Model + few shot learning)** | 13.09 | 1.95** | 0.47* | 12.40 | 2.86** | 1.28* | |
| cosα, speed + enhanced DTW [7] | - | - | 3.06 | - | - | - | 10 |
| pole-zero models [22] | - | 2.09 | - | - | - | - | 05 |
| DCT and sparse representation [22] | - | - | 0.51 | - | - | - | 10 |
| with all domain [23] | - | - | 3.88 | - | - | - | 10 |
| with stable domain [23] | - | - | 2.13 | - | - | - | 10 |
| Kinematic Theory of rapid human movements[24] | 7.87 | - | - | 3.61 | - | - | 01 |
| writer dependent features and classifiers[45] | - | - | 1.92 | - | - | - | 10 |
| Length Normalization + Fractional Distance [45] | - | - | 3.52 | - | - | - | 10 |
| Target-Wise [29] | 6.67* | - | - | 1.55* | - | - | 10 |
| Stroke-Wise [29] | 7.74** | - | - | 2.23** | - | - | 10 |
| Information Divergence-Based Matching [30] | - | 1.6* | 2.13 | - | - | - | 10 |
| Association of curvature feature with Hausdorff distance [38] | - | 7.05 | - | - | 1.02* | - | |

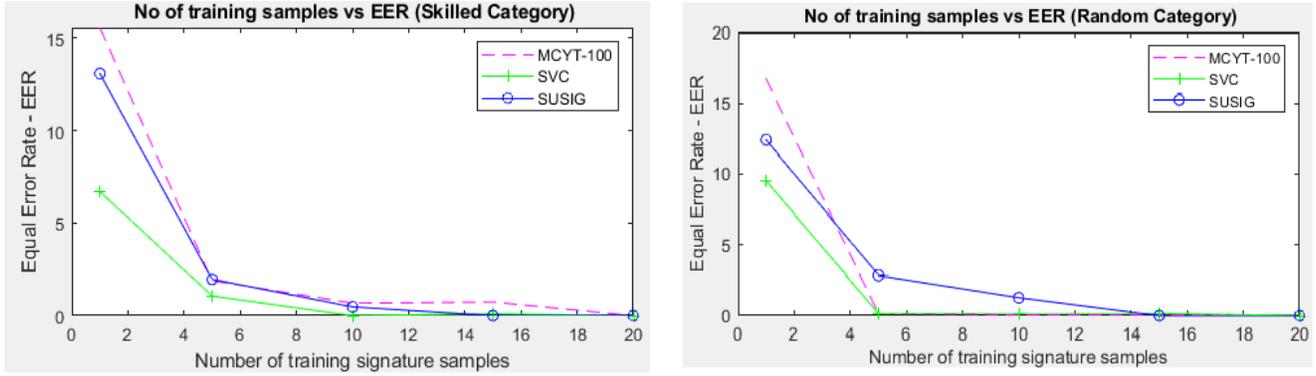

Figure. 2. The average EER with three different datasets for (a) Skilled Forgeries and (b) Random Forgeries.

## 4. Experimentation and results

In this segment, we present the experimental set up, results and performance analysis with the state of the art and recent literature. We have conducted the experiments on an Ubuntu machine containing Ubuntu 16.04 LTS, Intel Core i7-7700 CPU, with Titan X GPU. The proposed models are implemented in keras using python with Tensorflow [28,41] backend. We have conducted experimentations on the MCYT-100 online signature dataset (DB1), Visual Subcorpus of SUSIG, SVC task 2004 database. The complete details of these datasets are illustrated in Table 1.

Similar to [7,9,20], the metrics used to evaluate the efficiency of the proposed system are: (i) False Rejection Rate (FRR), which signifies the percentage of genuine signatures that are rejected by the system, (ii) False Acceptance Rate (FAR), characterizes the percentage of forgeries that are accepted (FAR can be computed for each of skilled and random forgeries), (iii) Average Error Rate (AER), is the average error considering FRR, FAR (random, skilled). (iv) Equal Error Rate (EER), is the point at which the FRR = FAR.

We have trained the system with varied number of genuine signatures 1,5,10,15,20 and with equal number of random forgery samples for each user. Genuine signatures of other writers are taken as a random forgery for a writer. Further, the training set is split into training and validation set. Sixty percent of the available training samples are used to fine tune the model. During fine tuning the model, the focus is on minimizing the Equal Error Rate (EER). Same set of hyper parameters (epochs, learning rate, batch size, number of CNN layers, number of LSTM layers, number of filters etc.) are used for testing also. We conducted experimentation for different number of trials 'T' and in each trial, training and testing signatures were randomly selected. In our work, with T = 20, best results were achieved.

As the main motive of the current work is for resource constraint devices/networks, we had further studied the effect of number of features considered for training the model. The EER of the proposed model of different datasets for varying number of features is shown in Table II, III and IV. Table II illustrates that in case of MCYT-100 dataset, if we consider 80 features, the EER (Skilled_10,15,20 and Random_10,15,20) shown better performance compared to almost all the models presented in the literature, which considered all the 100 features. In case of SVC-2004 dataset, as depicted in Table III, the EER resulted by considering 40 features achieved the state-of-the art performance in categories of (Skilled_01,05,10,15 and Random_10, 15). In case of SUSIG, as depicted in Table IV, the EER resulted by considering 40 features achieved the state-of-the art performance in categories of (Skilled_05, 10 and Random_5, 10). As illustrated in table II,III,IVI, even though the frameworks proposed in [32,37,38] are resulting in better EER values compared to the proposed framework, these models are evaluated with categories of skilled_1, random_1, whereas we have extensively evaluated the model with all the possible training samples (1,5,10,15,20). Hence, its superiority proved compared to [32,37,38].

Figure 2, illustrates that our proposed model converges to zero EER as the number of training signature sample increases. In case of skilled category, SVC dataset shows faster convergence and in case of random category MCYT and SVC converges to zero EER.

## 5. CONCLUSION

The main contribution of this work is to develop OSV models for resource constraint mobile devices. Based on the author's knowledge, this work provides the first, complete and successful framework on the use of CNN and LSTM combination for OSV with reduced feature set. The main advantage of the proposed model is that it achieves few shot learning in which the framework learns the user specific features with one/few signature samples and achieves state of the art results in S_10,15,20,R_10,15,20 categories of MCYT, all categories of SVC datasets except R_01,R_05. S_05,10,R_05,10 categories of SUSIG datasets and reflects the realistic scenario. The proposed model is thoroughly tested on multiple datasets and proved its efficiency with reduced error rates. In future work, we are planning to work on models for OSV with mobile device interoperability (combination of different writing tools, acquiring the training and testing signatures from users on various devices) and further reduced feature set.